\def\year{2020}\relax
\documentclass[letterpaper]{article} % DO NOT CHANGE THIS
\usepackage{aaai20}  % DO NOT CHANGE THIS
\usepackage{times}  % DO NOT CHANGE THIS
\usepackage{helvet} % DO NOT CHANGE THIS
\usepackage{courier}  % DO NOT CHANGE THIS
\usepackage[hyphens]{url}  % DO NOT CHANGE THIS
\usepackage{graphicx} % DO NOT CHANGE THIS
\usepackage{graphicx}  % DO NOT CHANGE THIS
\usepackage{amsmath}
\usepackage{amsfonts}
\usepackage{booktabs}
\usepackage{subfigure}
\usepackage{algorithm}
\usepackage{algorithmic}

\newcommand{\relu}{\text{relu}}
\newcommand{\hA}{\hat{A}}

\newcommand{\he}{\bar{e}}
\newcommand{\B}{\mathcal{B}}
\newcommand{\N}{\mathcal{N}}

\newcommand{\D}{\Delta}
\newcommand{\KL}{D_{\mathrm{KL}}}

\title{Effective Decoding in Graph Auto-Encoder Using Triadic Closure}
\author{
    Han Shi,\textsuperscript{\rm 1}
    Haozheng Fan,\textsuperscript{\rm 1,2}
    James T. Kwok\textsuperscript{\rm 1}
    \\
    \textsuperscript{\rm 1}Department of Computer Science and Engineering\\
	  Hong Kong University of Science and Technology, Hong Kong\\
    \textsuperscript{\rm 2}Amazon\\
    \{hshiac,jamesk\}@cse.ust.hk, fanhaozh@amazon.com
}
\begin{document}

\maketitle

\begin{abstract}
The (variational) graph auto-encoder and its variants have been popularly used for representation learning on graph-structured data. While the encoder is often a powerful graph convolutional network, the decoder reconstructs the graph structure by only considering two nodes at a time, thus ignoring possible interactions among edges. On the other hand, structured prediction, which considers the whole graph simultaneously, is computationally expensive. In this paper, we utilize the well-known {\em triadic closure} property which is exhibited in many real-world networks. We propose the triad decoder, which considers and predicts the three edges involved in a local triad together. The triad decoder can be readily used in any graph-based auto-encoder. In particular, we incorporate this to the (variational) graph auto-encoder. Experiments on link prediction, node clustering and graph generation show that the use of triads leads to more accurate prediction, clustering and better preservation of the graph characteristics.
\end{abstract}

%%%%%%%%%%%%%%%%%%%%%%%%%%%%%%%%%%%%%%%%%%%%%%%%%%%%%%%%%%%%%%%%%%%%%%%%%%%%%%%%%%%

\section{Introduction}

With the proliferation of online social networks, an enormous number of people are now connected digitally.
Besides people, almost everything is also increasingly connected either physically or by all sorts of
relationships, leading to the recent popularity of the internet of things and knowledge graphs. In today's big data era, it is thus
important for organizations to derive the most value and insight out of this colossal amount of data entities and inter-relationships.

%A basic way to structure and analyze such a myriad of connections is by using graphs.

In general, nodes in the graph represent data entities, while edges
represent all sorts of fine-grained relationships. For example, in a social
network, the nodes are users that are connected by edges denoting pairwise
friendships. In an author collaboration network, the edges denote co-authorship
relationships.
Besides social networks and knowledge graphs, data in domains such as chemistry and natural language semantics are often naturally represented as graphs.

There are a number of important tasks in graph analytics.
A prominent example is
link prediction \cite{wang2017predictive},
which predicts whether an edge should exist between two given nodes.
Since its early success at LinkedIn, link recommendation has attracted significant
attention in
social
networks,  and also
in predicting associations between molecules in biology, and the discovery of relationships in a terrorist network.
Other popular graph analytics tasks include the
clustering
of nodes
\cite{wang2017mgae}, and
automatic
graph
generation
\cite{bojchevski2018netgan}.
Node clustering aims to partition graph nodes into a set of clusters such that the
intra-cluster nodes are much related (densely connected) with each other as compared
to the inter-cluster nodes. These cluster structures occur frequently in many
domains such as
computer networks, sociology, and physics. Graph generation refers to the task
of generating similar output graphs given an input graph. It can be used in the
discovery of molecule structures and dynamic network prediction.

However, graphs are typically difficult to analyze because they are large and
highly sparse.
Recently, there is a surge of interest in learning better
graph representations
\cite{goyal2018graph,hamilton2017representation}.
Different approaches are proposed to embed
the structural and attribute information in a graph
to a low-dimensional vector space,
such that
both the neighborhood similarity and community membership
are preserved
\cite{zhang2018network}.
A particularly successful unsupervised representation learning model on graphs
is the variational auto-encoder (VAE) \cite{kingma2013auto}, and its variants
such as the
variational graph auto-encoder
(VGAE),
graph auto-encoder
(GAE) \cite{kipf2016variational}, adversarially regularized graph auto-encoder
(ARGA),
and
adversarially regularized variational graph auto-encoder
(ARVGA)
\cite{pan2018adversarially}.
All these
auto-encoder
models consist of an encoder, which
learns the latent representation,
and a decoder, which reconstructs the graph-structured data based on the learned representation.
The encoder
is often based on the powerful
graph convolutional network (GCN) \cite{kipf2016semi}. However,
the decoder is relatively primitive, and prediction of a link is based simply on the
inner product between
the latent representations of the
nodes involved.

As nodes in a graph are interrelated, instead of  predicting each link separately,
all the candidate links in the whole
graph should be considered together, leading to a
structured prediction problem. However, learning of structured prediction models is often NP-hard \cite{globerson2015hard}. A recent model along this direction is
the structured prediction energy network (SPEN) \cite{belanger2016structured}.
However, design of the underlying energy function is still an open question, and
computation of the underlying Hessian is computationally expensive.

On the other hand,
an interesting property
exhibited
in many real-world networks
is the so-called
triadic closure, which is first proposed by
\citeauthor{simmel1908sociology}
(\citeyear{simmel1908sociology}).
This mechanism states that for any
three nodes
$\{i, j, k\}$ in a graph, if
there are edges between $(i,j)$ and $(i,k)$, it is
likely that an edge also exists between $j$ and $k$.
It is then popularized by
\citeauthor{granovetter1973strength}
(\citeyear{granovetter1973strength}), who
demonstrated empirically that triadic closure can be used to characterize link connections in many domains.
For example,
if two people (A and B)
in a friendship network
have a common friend, it is likely
that A and B are also friends.
If two papers
in a citation network
cite the same paper (suggesting that they belong to the same topic),
it is likely that one also cites the other.
Besides, the triadic closure  is also fundamental to the understanding and prediction of network evolution and
community growth
\cite{caplow1968two,bianconi2014triadic,zhou2018dynamic}.
%All above works focus on network growth and cannot be applied in static networks directly.

In this paper,
we utilize this triadic closure property as an efficient tradeoff
between
structured prediction (which
considers the whole graph simultaneously but is
expensive) and
individual link prediction  (which is simple but ignores interactions among
edges).
Specifically,
we propose the {\em triad decoder}, which
predicts the three edges involved
in a triad together.
The triad decoder
can readily replace the vanilla decoder in any graph-based auto-encoder.
In particular, we incorporate this into
VGAE
and
GAE,
leading to the {\em triad variational graph auto-encoder\/} (TVGA) and
{\em triad graph auto-encoder\/} (TGA).
Experiments are performed on link prediction, node clustering and graph generation
using a number of real-world data sets.
The prediction and clustering results are more accurate, and the graphs generated preserve more
characteristics of the input graph,
demonstrating the usefulness of triads in graph analytics.

%%%%%%%%%%%%%%%%%%%%%%%%%%%%%%%%%%%%%%%%%%%%%%%%%%%%%%%%%%%%%%%%%%%%%%%%%%%%%%%%%%%

\section{Related Work: Graph Auto-Encoder}
\label{sec:review}

The variational graph auto-encoder (VGAE) and its non-variational variant
graph auto-encoder
(GAE) are
introduced in
\cite{kipf2016variational}.
Using the
variational auto-encoder (VAE) framework \cite{kingma2013auto}, they are
based on unsupervised deep learning model
consisting of
an encoder and a decoder.

\subsection{GCN Encoder}

Let the graph be $G = (V, E)$, where $V$
is a set of $N$ nodes, and $E$ is the set of edges.
Let its adjacency matrix
be $A$.
The encoder is a graph convolutional network (GCN) \cite{kipf2016semi},
which is a deep learning model for graph-structure data.
For a $L$-layer GCN,
the layer-wise propagation rule is given by:

\[ H^{(l+1)}
=f(H^{(l)}, A)
=\relu(\widetilde{D}^{\frac{1}{2}}\widetilde{A}\widetilde{D}^{-\frac{1}{2}}H^{(l)}W^{(l)}), \]
where $\widetilde{A}=A+I$, $I$ is the identity matrix, $\widetilde{D}$ is a diagonal matrix with $\widetilde{D}_{ii}=\sum_{j=1}^NA_{ij}$,
$H^{(l)}$ and
$W^{(l)}$ are the
feature map
and weight
at the $l$th layer, respectively,
and $\text{relu}(\cdot)$ is the ReLU activation
function.
$H^{(0)}$ is the matrix of node feature vectors $X$,
and $H^{(L)}$ is the
(deterministic) graph embedding matrix $Z$.
Often, $L=2$
\cite{kipf2016variational},
leading to the following encoder:
\begin{align}
     H^{(1)} = f(X, A),\;\; Z = f(H^{(1)}, A). \label{eq:non-prob}
\end{align}

In VGAE, the embedding (encoder output) is probabilistic.
Let $z_i$ be the embedding of node $i$. It is assumed to follow the normal
distribution:
\begin{equation} \label{eq:encoder}
q(z_i|X, A) = \mathcal{N}(\mu_i, \text{diag}(\sigma^2)),
\end{equation}
where $\mu_i$ and $\log\sigma$ are outputs from two GCNs that share the first-layer weights.
The distribution for
all the embedding vectors is then
$q(Z|X, A) = \prod_{i=1}^N q(z_i|X, A)$.

%%%%%%%%%%%%%%%%%%%%%%%%%%%%%%%%%%%%%%%

\subsection{Inner Product Decoder }

The decoder is often
a vanilla model based on the
inner product between the latent representations of two nodes.
For GAE, the graph adjacency matrix
$\hat{A}$ is reconstructed from
the inner product of
two node
embeddings as:
\begin{equation} \label{eq:vanilladecoder}
   \hat{A}=\sigma(ZZ^T),
\end{equation}
where $\sigma$ is the sigmoid activation function.
For the VGAE, the decoder is also based on
inner products, but is probabilistic:
\[ p(\hat{A}|Z)=\prod_{i=1}^N\prod_{j=1}^N p(\hat{A}_{ij}|z_i,z_j), \]
where
$p(\hat{A}_{ij}|z_i,z_j)=\sigma(z_i^Tz_j)$.

%%%%%%%%%%%%%%%%%%%%%%%%%%%%%%%%%%%%%%%%%%%%%%%%%%%%%%%%%%%%%%%%%%%%%%%%%%%%%%%%%%%

\section{Triad Decoder}
\label{sec:triad}

% Let the probability that nodes $i$ and $j$  are connected be $e_{ij}$.

Based on triadic closure,
the presence of a particular edge in a triad is dependent on whether the other two edges are present.
Specifically,
consider the triad $\D=(i, j, k)$.
Let $I_{ij}$ be the indicator function representing whether nodes $i$ and $j$ are
connected (i.e., $I_{ij}=1$, if $i, j$ are connected; and 0 otherwise).
As the presence of  edges
$(i,j),
(i,k)$, and
$(j,k)$ are interrelated,
we propose to predict the three edge probabilities
$\{P(I_{ij}|\D), P(I_{ik}|\D), P(I_{jk}|\D)\}$ (denoted
$\{e_{ij}(\D), e_{ik}(\D), e_{jk}(\D)\}$) together,
using as inputs
the three embeddings $z_i, z_j, z_k$, where $z_i$'s are the
latent representations of the nodes learned by the graph encoder.

%%%%%%%%%%%%%%%%%%%%%%%%%%%%%%%%%%%%%%%

\subsection{Structure}

\begin{figure}
\centering
\includegraphics[width=.7\linewidth]{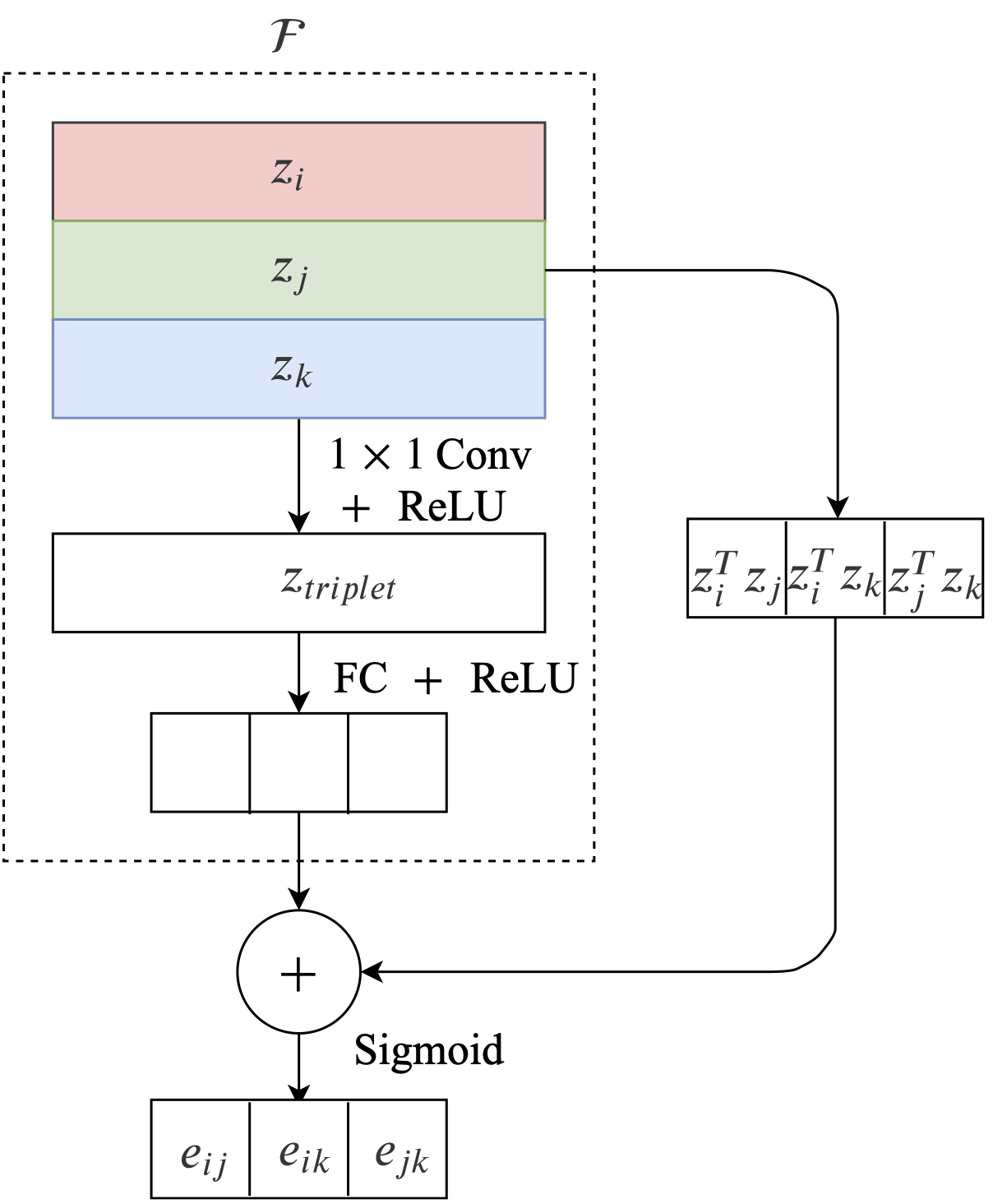}
\vspace{-.1in}
\caption{Structure of the triad decoder.}
  \label{fig:decoder}
\end{figure}

The structure of the proposed triad decoder is
shown in Figure~\ref{fig:decoder}.
First, we perform $1\times 1$ convolution,
which corresponds to a linear transform,
on the three node embeddings $z_i,z_j,z_k$.
This is followed by the rectified linear (ReLU) nonlinearity, producing
the vector $z_{\text{triplet}}$. Its output for the
$l$th
dimension
is $\text{ReLU}(w_i z_{il}+w_jz_{jl}+w_kz_{kl})$. Note that this
contains information from the whole triad.
Vector $z_{\text{triplet}}$
is then further nonlinearly transformed by the fully connected (FC) layer and
another ReLU nonlinearity. The whole block, which is denoted $\mathcal{F}$, is
finally merged with the three inner
products constructed from $z_i,z_j,z_k$.  This additional inner product connections serve similarly as the
residual connection in residual networks \cite{he2016deep}.
The output of the triad decoder is:
\begin{eqnarray}
\lefteqn{[e_{ij}(\D), e_{ik}(\D), e_{jk}(\D)]} \nonumber\\
& = & \sigma(\mathcal{F}(z_i, z_j, z_k) + [z_i^Tz_j, z_i^Tz_k, z_j^Tz_k]). \label{eq:dec1}
\end{eqnarray}
When  $\mathcal{F}$ outputs a zero mapping,
it reduces to the standard inner product decoder.

%The practical meaning of block $\mathcal{F}$ is the effect of the third node to the pair nodes, while inner product is the pair-wise effect.

%%%%%%%%%%%%%%%%%%%%%%%%%%%%%%%%%%%%%%%

\subsection{Model Training}
\label{sec:train}

\begin{figure*}[th]
\centering
\includegraphics[width=.8\linewidth]{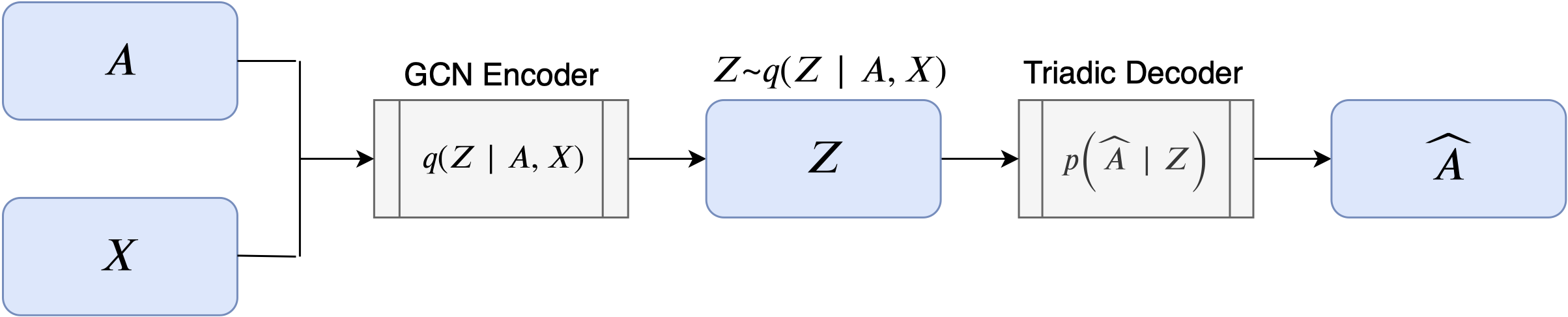}
\vspace{-.1in}
\caption{Structure of the graph-based auto-encoder
with a triad decoder.}
  \label{fig:diag}
\end{figure*}

The triad decoder can be readily used to replace the decoder in any graph-based auto-encoder. In this paper, we
incorporate this into the VGAE and GAE, leading to
the {\em triad variational graph auto-encoder\/} (TVGA)
and
the {\em triad graph auto-encoder\/} (TGA), respectively
(Figure~\ref{fig:diag}).

The whole network can be trained end-to-end, and
the training procedure is shown in Algorithm \ref{alg:algorithm}.
The
total number of triads
in the graph
is ${N \choose 3}$,
which is large.
As is common in deep learning, we
use stochastic gradient descent (SGD) or its variants for better scalability. In particular,
the popular
Adam optimizer \cite{kingma2014adam} will be employed
in the experiments.
In each iteration, we sample $B$ triads
$\{\D_1,\dots,\D_B\}$
from the graph to form a mini-batch $\B$.
The probability for the presence of a particular edge $(i,j)$ in the mini-batch
is
\begin{equation} \label{eq:ave}
\he_{ij}=E_{ij}/M_{ij},
\end{equation}
where $E_{ij}=\sum_{m=1}^B e_{ij}(\Delta_m)$, with
$e_{ij}(\Delta_m)=0$ if $(i,j)\notin\Delta_m$.
%(i.e., $E_{ij}=\sum_{m=1}^B e_{ij}(\Delta_m)$, with $e_{ij}(\Delta_m)=1$ if $(i,j)\in\Delta_m$, and 0 otherwise),
$M_{ij}$
is
the total number of times $(i,j)$ observed in the mini-batch
(i.e., $M_{ij}=\sum_{m=1}^B I_{ij}(\Delta_m)$,
where
$I_{ij}(\Delta_m) =1$ if $(i,j)\in\Delta_m$; and 0 otherwise).

\begin{algorithm}[htb]
\caption{Training the triad variational graph auto-encoder (TVGA) and
triad graph auto-encoder (TGA) using
SGD. Here,
the encoder and decoder parameters are combined and denoted by $w$.
$L_w(\{\he_{ij}\}, A, X)$ is the loss function (for TVGA,
it is the negative of (\ref{eq:loss_vae});
for TGA,
it is (\ref{eq:loss_ae})).}
\label{alg:algorithm}
\begin{algorithmic}[1] %[1] enables line numbers
\STATE initialize $w_0$;
\STATE $t\leftarrow 0$;
\WHILE{$w$ not converged}
\STATE $t\leftarrow t+1$;
\STATE encode the graph $G$ to latent embedding $Z$ using (\ref{eq:non-prob}) for
TGA and (\ref{eq:encoder}) for TVGA;
\STATE sample $B$ triads to form mini-batch $\B$;
\STATE obtain $\he_{ij}$'s for the triads from
(\ref{eq:dec1}) and
(\ref{eq:ave});
\STATE $w_t\leftarrow w_{t-1}-\alpha\nabla_wL_{w_{t-1}}(\{\he_{ij}\}, A, X)$;
\ENDWHILE
\STATE \textbf{return} $Z$.
\end{algorithmic}
\end{algorithm}

%%%%%%%%%%%%%%

\subsubsection{Optimization Objective}

First, we consider TVGA.
Let $A_{ij}$ be the observed graph adjacency matrix.
As in \cite{kipf2016variational},
this variational model
is trained
by maximizing the variational lower bound:
\begin{align}
& \sum_{\text{triad} \in \B}
\sum_{(i,j)\in \text{triad}} (A_{ij}\log \he_{ij}+(1-A_{ij})\log (1-\he_{ij})) \nonumber\\
& -\KL[q(Z|X, A)||p(Z)] \label{eq:loss_vae}
\end{align}
w.r.t. the encoder
parameters
($W^{(1)}, W^{(2)}$ of the GCN)
and decoder parameters.
Here, the
first term is the negative of the graph reconstruction error, while the second term
is the Kullback-Leibler divergence between the encoder output distribution for the
embeddings ($q(\cdot)$ in (\ref{eq:encoder})) and some prior distribution
$p(\cdot)$. In this paper,
we
assume that the latent dimensions are i.i.d., and follow the standard normal distribution: $p(Z)=\prod_{i=1}^N p(z_i)=\prod_{i=1}^N\mathcal{N}(0, I)$.

As for the non-variational version
TGA,
the Kullback-Leibler divergence
is removed
from (\ref{eq:loss_vae})
as in \cite{pan2018adversarially},
and only the following
graph reconstruction error is
minimized:
\begin{equation} \label{eq:loss_ae}
-\sum_{\text{triad} \in \B}
\sum_{(i,j)\in \text{triad}} (A_{ij}\log \he_{ij}+(1-A_{ij})\log (1-\he_{ij})).
\end{equation}

%%%%%%%%%%%%%%

\subsubsection{Sampling the Training Triads}

Real-world networks are typically sparse.
If triads are sampled randomly
from the graph
in step~6 (of
Algorithm~\ref{alg:algorithm}),
it is likely that most node pairs
in the training triads are disconnected.  This resultant high class imbalance can lead
to severe performance deterioration for most classifiers.

To alleviate this problem, we construct a more balanced training data set by
sampling each
triad $\D=(i, j, k)$ as follows.
First,
a node $i$ is
randomly sampled from the graph.
Let
$\N(i)$
be
the set
containing all
neighbors of $i$.
With probability $p$,
we sample the next node $j$ from $\N(i)$;
and with probability $1-p$,
sample
$j$ from a faraway node not in $\N(i)$.
After sampling $j$,
the last node $k$ in the triad
is
similarly
sampled
from $\N(j)$
with probability $p$;
and from a node not in $\N(j)$ with probability
$1-p$.

The class imbalance can be controlled by appropriately setting $p$. First,
note that
the expected number of connected edges
in a triad $\D$
is $E[I_{ij}+I_{jk}+I_{ik}] = E[I_{ij}]+E[I_{jk}]+E[I_{ik}]$.
Using the above sampling scheme,
$E[I_{ij}]=E[I_{jk}]=p$.
As for $E[I_{ik}]$, this depends on the cases where
pairs $(i, j)$ and/or $(j, k)$ are connected (which are independent
based on the sampling scheme). Thus,
% \begin{eqnarray*}
% \lefteqn {E[I_{ij}+I_{jk}+I_{ik}]} \\
% & = & p + p +
% p^2 \cdot P(I_{ik}| \text{observed 2 edges}) \\
% && + 2p(1-p) \cdot P(I_{ik}| \text{observed 1 edge}) \\
% && + (1-p)^2 \cdot
% P(I_{ik}| \text{observed 0 edge}).
% \end{eqnarray*}
\begin{align*}
E[I_{ij}+I_{jk}+I_{ik}] = & p + p +
p^2P(I_{ik}| \text{observed 2 edges}) \\
& +2p(1-p)  P(I_{ik}| \text{observed 1 edge}) \\
& + (1-p)^2
P(I_{ik}| \text{observed 0 edge}).
\end{align*}
$P(I_{ik}| \text{observed 2 edges})$
can be estimated by
the global clustering coefficient
\cite{wasserman1994social},
which is defined as the ratio of
the number of closed triads (i.e., all three nodes in the triad are connected  to each other)
to the total number of triads.
For $P(I_{ik}| \text{observed 1 edge})$ and
$P(I_{ik}| \text{observed 0 edge})$,
intuitively, the effect of the presence of one or zero edge on
$I_{ik}$ is small. Thus, we
simply assume that both probabilities are the same as the prior probability $P(I_{ik})$,
which can be estimated from the graph density
\cite{lawler2001combinatorial}.
Hence,
\begin{eqnarray*}
E[I_{ij}+I_{jk}+I_{ik}]
& = & 2p + p^2 \cdot \text{(clustering coefficient)} \\
&& +(1-p^2) P(I_{ik}).
\end{eqnarray*}
To construct a balanced training set,
the desired $p$
can be obtained
by setting
the above to $3/2$.

%%%%%%%%%%%%%%

\subsubsection{Space Complexity}

VGAE/GAE
\cite{kipf2016variational}
and
ARVGA/ARGA
\cite{pan2018adversarially}
take the whole graph as input.
The space complexities are both $\mathcal{O}(N^2)$, where $N$ is the number of nodes.
For large graphs, the adjacency matrix may not even be able to fit into memory.
In contrast, the proposed
algorithm
is trained
on mini-batches of triads.
The space complexity is $\mathcal{O}(\max(N,
B))$,
which are much smaller than $N^2$ and thus much more scalable.

%%%%%%%%%%%%%%%%%%%%%%%%%%%%%%%%%%%%%%%

\section{Inference}

After training, the learned model can be used on a variety of graph learning tasks.
In this section, we focus on
link prediction, node clustering and graph generation.

\subsection{Link Prediction}

In link prediction, one wants to predict whether an edge exists between nodes $i$ and $j$.
Recall that the proposed triad decoder predicts all three edges in the whole triad
simultaneously, and so the three nodes need to be inputted together.
In constructing these triads during inference, intuitively
a node faraway from the node pair $(i,j)$ carries little information.
Hence,
instead of
using both neighboring and faraway nodes to construct triads as in
training,
we only aggregate predictions from
nodes $k$ that are
in $\N(i)\cup \N(j)$.
For each  such $k$, the decoder predicts
the probability
for each $(i,j)$ edge
using (\ref{eq:dec1}).
The average probability over all these $k$'s
is taken
as the final
probability
for the presence of an edge.

On using
NetGAN for link prediction, one has to first
generate a number of random walks
and
accumulate the corresponding transition counts
to an $N\times N$ matrix, which is then thresholded to produce the predicted links.
Thus, the space and time
complexities
are both $\mathcal{O}(N^2)$.
In contrast,
the proposed algorithm predicts
the edge
directly by (\ref{eq:ave}) in
$\mathcal{O}(1)$ time.
With $N_{\text{neighbr}}$ nodes in the union of neighborhoods of $i$ and $j$ (typically,
$N_{\text{neighbr}} \ll N$),
the total complexity is
$\mathcal{O}(N_{\text{neighbr}})$.
Hence, the proposed algorithm is much more efficient.
On the other hand,
VGAE/GAE and ARVGA/ARGA
only take $\mathcal{O}(1)$ space and time. However, as will be
seen in the experiments,
their link prediction results are much inferior.

%%%%%%%%%%%%%%
\subsection{Node Clustering}

With the learned node embedding,
one can apply a standard clustering algorithm
to
cluster the nodes.
Recall that the encoder and decoder
in a graph-based auto-encoder
are trained together in an end-to-end manner.
Hence, though the decoder is not explicitly used in
node clustering, an improved decoder (such as the proposed triad decoder)  can guide the learning of better node
representations in the embedding space.

\subsection{Graph Generation}

TVGA, which is based on
the VAE framework, can be used to generate graphs. TGA, on the other hand, is not probabilistic
and cannot be used for
graph generation.

Let $N'$ be the target number of nodes to be generated.
We first randomly sample $N'$ $z_i$'s
from the posterior (normal) distribution of TVGA in the latent space.
We
then
randomly sample $K$ triads
from these $z_i$'s.
For each triad,
(\ref{eq:dec1})
is used
to predict the probabilities for the three constituent edges.
Finally, the predictions are averaged over
all $K$ triads, and
stored in an estimated adjacency matrix $\hat{A}$.
To ensure symmetry, we replace each entry
$\hat{A}_{ij}$ of $\hat{A}$ by
$(\hat{A}_{ij}+\hat{A}_{ji})/2$.

We assume that the generated graph has to be connected.  Inspired by NetGAN, the
following strategy is used to generate such a graph
from $\hA$.
First,
for every node $i$,
we sample an edge $e_{ij}$ to node $j$
with probability $p_{ij} = \hat{A}_{ij}/\sum_{k=1}^N \hat{A}_{ik}$, and add it to the graph if it is new.
%If $e_{ij}$ has been sampled before (i.e., connectivity has been ensured), we ignore the second sampling.
Afterwards,
we continue adding edges to the graph
until the target number of edges have been generated.
However, unlike NetGAN which uses sampling,
we simply add edges
in
descending probability $p_{ij} = \hat{A}_{ij}/\sum_{u=1}^N\sum_{v=1}^N \hat{A}_{uv}$.
Empirically, this has better performance
as the whole local triad information is used in the proposed algorithm, and so
$p_{ij}$ is more reliable.

%%%%%%%%%%%%%%%%%%%%%%%%%%%%%%%%%%%%%%%%%%%%%%%%%%%%%%%%%%%%%%%%%%%%%%%%%%%%%%%%%%%

\section{Experiments}
\label{sec:expt}

In this section, we demonstrate the performance of the proposed algorithm on
link prediction,
node clustering and graph generation.
Experiments are performed on
three standard benchmark
citation
graph data
sets\footnote{\url{http://www.cs.umd.edu/~sen/lbc-proj/LBC.html}}
\cite{sen2008collective}: Cora, Citeseer, and Pubmed
(Table~\ref{tab:graph}).
Each node
represents an article,
and has a boolean feature vector whose entries indicate whether a specific word
occurs in the article. Each node also has a label, indicating the class it belongs to.
The edges are citation links.
We treat the graphs as undirected graphs,
and all self-loops are removed.
Moreover, we only use the largest connected component in each graph.

\begin{table}[t]
\centering
\caption{Statistics for the (largest connected component of) graph data sets used.}
\begin{tabular}{cccc}
\toprule
& Cora & Citeseer & PubMed  \\
\midrule
number of nodes     & 2,485 & 2,120 & 19,717  \\
number of edges & 5,069 & 3,679 & 44,324   \\
number of classes   & 7 & 6 & 3   \\
feature dim & 1,433 & 3,703 & 500 \\
clustering coefficient & 0.2376 & 0.1696 & 0.0602\\
graph density & 0.0016 & 0.0016 & 0.0002 \\
\bottomrule
\end{tabular}
\label{tab:graph}
\end{table}

%%%%%%%%%%%%%%%%%%%%%%%%%%%%%%%%%%%%%%%
\subsection{Link Prediction}
\label{sec:link}

In this experiment,
85\% of the edges
and non-edges (unconnected nodes)
from each graph
are randomly selected to form the training set, another
10\% is used as the validation set, and the remaining 5\%
as testing set.
The proposed algorithm uses a mini-batch size of
5,000.
Adam
\cite{kingma2014adam} is the optimizer,
with a learning rate
of $0.0005$.
Both the hidden layer and embedding layer
of the encoder
have 32 hidden units.
The convolution layer in the triad decoder has 4 filters (i.e., the dimension of
$z_{\text{triplet}}$ is $1\times 32\times 4$).

\subsubsection{Triad Sampling Scheme}
\label{sec:sampling}

\begin{figure*}[htbp]
\begin{center}
\includegraphics[width=1.8\columnwidth]{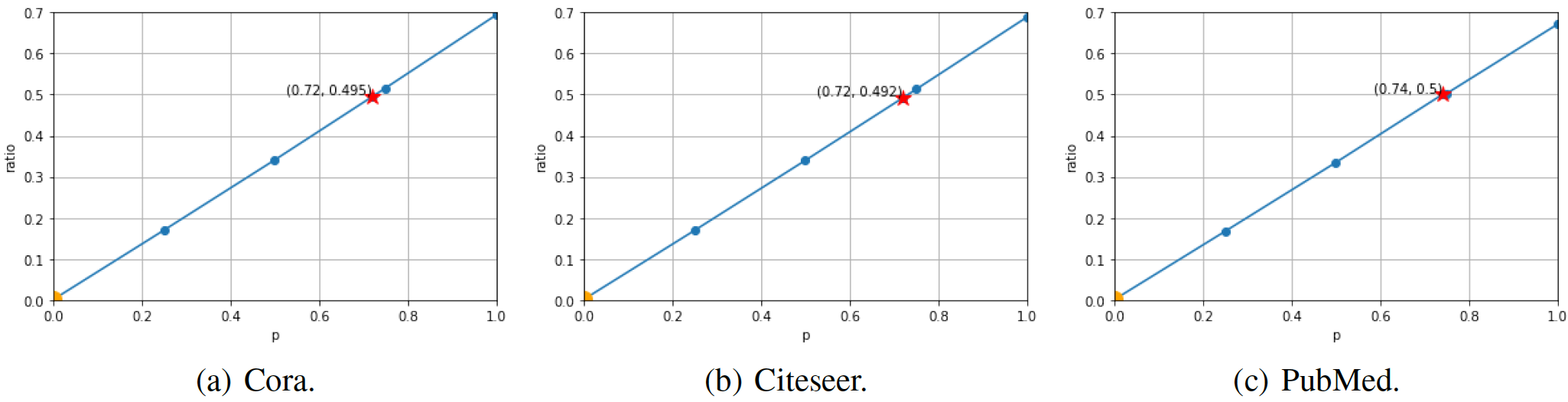}
\vspace{-.1in}
\caption{Ratio of existent edges in each triad, with different sampling probabilities $p$.}
\label{fig:balance}
\end{center}
\end{figure*}

First, we study how
the proposed sampling scheme alleviates the class imbalance problem.
Figure~\ref{fig:balance} shows the proportion of existent edges in the triads,
with different sampling probabilities $p$.

With random sampling (yellow dot in the figure), the ratio of existent edges is
very small,
and
the data set
is highly imbalanced.
As expected,
using a larger $p$ means
that node $j$ is more likely to be connected to node $i$, and
node $k$ is more likely to be connected to node $j$, leading to a higher proportion
of edges being observed in the sampled triads.
The red dot corresponds to the $p$ value obtained by the proposed sampling scheme.
As can be seen, the ratios of existent edges are all close to $0.5$ on the three data sets,
indicating that the proposed sampling scheme has effectively alleviated the
class imbalance problem.

\subsubsection{Comparison with the State-of-the-Art}

The proposed algorithm (using both random triad sampling  and the proposed sampling
scheme) is
compared with the following baselines:
(i) node2vec
\cite{grover2016node2vec};
(ii) spectral clustering
(SC) \cite{bruna2013spectral}; (iii) NetGAN\footnote{NetGAN has two early stopping schemes. Here, we use VAL-CRITERION as
it is more related to generalization properties.}
\cite{bojchevski2018netgan}; (iv) SEAL \cite{zhang2018link}; (v) variational graph
auto-encoder (VGAE) \cite{kipf2016variational}; (vi) graph auto-encoder (GAE)
\cite{kipf2016variational}; (vii) adversarially regularized variational graph
auto-encoder (ARVGA) \cite{pan2018adversarially}; and (viii) adversarially
regularized graph auto-encoder (ARGA) \cite{pan2018adversarially}.
SC and node2vec
are only used
to generate node embeddings. The inner product decoder
in (\ref{eq:vanilladecoder}) is then used to obtain edge probabilities.
NetGAN generates random walks and accumulates the corresponding transition counts,
which are used to measure how likely there is an edge between two nodes.
Note that node2vec, SC and NetGAN do not make use of node features.
The other four baselines
use the graph convolutional network \cite{kipf2016semi} as encoder.
In particular,
ARVGA and ARGA are modified from
VGAE  and
GAE, respectively,
by adding a discriminator that are trained adversarially.

For performance evaluation, we use the area under
the ROC curve (AUC) and average precision (AP)
as in \cite{kipf2016variational}.

Results are shown in Table \ref{tab:linkprediction}.
As can be seen,
node2vec,
SC and
NetGAN,
which do not utilize node features, have the worst performance.
TVGA and TGA outperform the other VAE-based methods, showing that triad information leads to more accurate
predictions.
Moreover, the proposed triad sampling scheme performs significantly better than
random triad
sampling (TVGA(rand) and
TGA(rand)).
Hence, we will only experiment with
the proposed triad sampling scheme in the sequel.

\begin{table}[t]
\centering
\caption{Link prediction accuracy (\%).
TVGA(rand) and TGA(rand)  are variants of
TVGA and TGA, respectively, that use random triad sampling.}
\resizebox{.95\columnwidth}{!}{
\begin{tabular}{ccccccc}
\toprule
& \multicolumn{2}{c}{Cora}  & \multicolumn{2}{c}{Citeseer} & \multicolumn{2}{c}{PubMed} \\
& AUC & AP & AUC & AP & AUC & AP \\
\midrule
node2vec & 86.9 & 88.7  & 88.1 & 89.2 & 92.2 & 92.3    \\
SC       & 87.8 & 91.5  & 86.2 & 89.2 & 97.1 & 96.1    \\
NetGAN   & 90.9 & 92.7  & 92.9 & 94.6 & 88.2 & 88.1    \\
SEAL     & 93.3 & 94.6  & 92.8 & 93.4 & 95.8 & 96.3    \\
VGAE     & 94.4 & 95.9  & 93.4 & 95.2 & 96.2 & 96.4 \\
GAE      & 93.1 & 95.0  & 91.5 & 92.6 & 97.1 & 97.4 \\
ARVGA    & 93.8 & 94.8 & 94.4 & 95.7 & 98.0 & 98.2  \\
ARGA     & 94.2 & 95.6  & 93.5 & 95.0 & 95.5 & 96.0 \\
\midrule
TVGA(rand)& 68.5 & 68.5 & 60.5 & 65.3 & 82.9 & 81.2 \\
TVGA & \textbf{96.0} & \textbf{96.3} & 96.2 & 96.5 & \textbf{98.5} & \textbf{98.7} \\
TGA(rand) & 76.3 & 76.9 & 69.2 & 69.1 & 89.6 & 88.7 \\
TGA & 95.6 & 96.2 & \textbf{96.5} & \textbf{97.0} & 98.2 & 98.4\\

\bottomrule
\end{tabular}
}
\label{tab:linkprediction}
\end{table}

%%%%%%%%%%%%%%%%%%%%%%%%%%%%%%%%%%%%%%%

\subsection{Node Clustering}
\label{sec:clus}

In this section, we consider the unsupervised task of clustering nodes in the graph.  We perform
clustering on the obtained node embeddings using the $K$-means clustering algorithm,
where $K$ is set to be the number of classes  in Table~\ref{tab:graph}.
The following baselines are compared with the proposed algorithm:
(i) node2vec;
(ii) spectral clustering (SC);
(iii) text-associated DeepWalk (TADW) \cite{yang2015network};
(iv) VGAE;
(v) GAE;
(vi) ARVGA;
and
(vii) ARGA.
SC is a general clustering algorithm, while TADW is designed specifically for graphs.
We do not compare with
SEAL and NetGAN, since they
do not produce node embeddings
for clustering.

The node labels in Table~\ref{tab:graph} are used as ground-truth clustering labels.
For performance evaluation,
we follow
\cite{xia2014robust}
and
first match the predicted
labels with the ground-truth labels using the Munkres assignment algorithm
\cite{munkres1957algorithms},
and then report the (i) accuracy (acc);
(ii) normalized mutual information (NMI); (iii)
F1-score (F1);
(iv)
precision;
and (v) adjusted rand index
(adj-RI).
As in \cite{pan2018adversarially}, we only evaluate on the Cora and Citeseer data sets.

Results
on Cora and Citeseer
are
shown in Tables~\ref{tab:coraclustering} and
\ref{tab:citeseerclustering}.
TVGA
and
TGA
outperform the other methods on both data sets across all metrics. SC does not perform well, as it does
not utilize node features and is not designed for graphs.

\begin{table}[t]
\caption{Node clustering performance on Cora.}
	\label{tab:coraclustering}
	\centering
  \resizebox{.95\columnwidth}{!}{
    \begin{tabular}{cccccc}
    \toprule
    & acc & NMI & F1 & precision & adj-RI \\
    \midrule
    node2vec  & 0.674 & 0.475 & 0.658 & 0.689 & 0.422        \\
    SC        & 0.299 & 0.077 & 0.088 & 0.294 & 0.001        \\
    TADW      & 0.604 & 0.438 & 0.553 & 0.613 & 0.320        \\
    VGAE      & 0.662 & 0.482 & 0.639 & 0.648 & 0.433        \\
    GAE       & 0.602 & 0.460 & 0.591 & 0.591 & 0.389        \\
    ARVGA     & 0.616 & 0.457 & 0.599 & 0.608 & 0.378        \\
    ARGA      & 0.687 & 0.518 & 0.677 & 0.692 & 0.455        \\
    \midrule
    TVGA & \textbf{0.753} & \textbf{0.591} & \textbf{0.731} & \textbf{0.781} & \textbf{0.560} \\
    TGA & 0.728 & 0.558 & 0.711 & 0.747 & 0.512\\
    \bottomrule
    \end{tabular}
    }
\end{table}

\begin{table}[t]
    \centering
    \caption{Node clustering performance on Citeseer.}
    \resizebox{.95\columnwidth}{!}{
    \begin{tabular}{cccccc}
    \toprule
    & acc & NMI & F1 & precision & adj-RI \\
    \midrule
    node2vec  & 0.478 & 0.291 & 0.461 & 0.531 & 0.235        \\
    SC        & 0.258 & 0.037 & 0.090 & 0.247 & 0.003        \\
    TADW      & 0.581 & 0.371 & 0.480 & 0.481 & 0.354
     \\
    VGAE      & 0.515 & 0.332 & 0.486 & 0.551 & 0.273        \\
    GAE       & 0.430 & 0.245 & 0.421 & 0.555 & 0.118        \\
    ARVGA     & 0.580 & 0.337 & 0.530 & 0.547 & 0.322        \\
    ARGA      & 0.584 & 0.370 & 0.536 & 0.572 & 0.339        \\
    \midrule
    TVGA & 0.591 & 0.365 & 0.544 & 0.565 & 0.339\\
    TGA & \textbf{0.611} & \textbf{0.401} & \textbf{0.564} & \textbf{0.600} & \textbf{0.387}\\
    \bottomrule
    \end{tabular}
    }
    \label{tab:citeseerclustering}
\end{table}

%%%%%%%%%%%%%%%%%%%%%%%%%%%%%%%%%%%%%%%

\subsection{Graph Generation}
\label{sec:gen}

\begin{table*}[t]
\centering
\caption{Statistics for the generated graphs on Cora. Results are averaged over 5 runs.}
\resizebox{0.95\textwidth}{!}{
\begin{tabular}{crrrrrrrrrrrrrr|c}
\toprule
& \multicolumn{2}{c}{Gini coeff.} & \multicolumn{2}{c}{max degree} &
\multicolumn{2}{c}{triangle count} & \multicolumn{2}{c}{assortativity} &
\multicolumn{2}{c}{power law exp.} & \multicolumn{2}{c}{clustering coeff.} &
\multicolumn{2}{c|}{charac. path len.} & \multicolumn{1}{c}{avg rank} \\
& \multicolumn{1}{c}{avg} & \multicolumn{1}{c}{std} & \multicolumn{1}{c}{avg} &
\multicolumn{1}{c}{std} & \multicolumn{1}{c}{avg} & \multicolumn{1}{c}{std} &
\multicolumn{1}{c}{avg} & \multicolumn{1}{c}{std} & \multicolumn{1}{c}{avg} &
\multicolumn{1}{c}{std} & \multicolumn{1}{c}{avg} & \multicolumn{1}{c}{std} &
\multicolumn{1}{c}{avg} & \multicolumn{1}{c|}{std} & \\
\midrule
original input & 0.397 &       & 168  &      & 1558  &      & -0.071  &       & 1.885 &       & 4.24e-3 &         & 6.31 &      & \\
\midrule
conf. model  & 0.397 & $\pm$ 0.000 & 168& $\pm$ 0.00& 113.6 &$\pm$ 10.8 & -0.019 &$\pm$ 0.008& 1.885 &$\pm$ 0.000 & 3.09e-4 &$\pm$ 2.94e-5 & 4.82 &$\pm$ 0.01 & 3.57 \\
DC-SBM       & 0.476 &$\pm$ 0.003 & 123 &$\pm$ 7.06 & 333 &$\pm$ 25.5 & -0.028  &$\pm$ 0.007 & 1.854 &$\pm$ 0.004 & 1.75e-3 &$\pm$ 1.15e-4 & 4.88 &$\pm$ 0.02 & 4.14 \\
NetGAN       & 0.372 &$\pm$ 0.003 & 131 &$\pm$ 2.87 & 874.0 &$\pm$ 13.5 & -0.074 &$\pm$ 0.003 & 1.861 &$\pm$ 0.002 & 4.63e-3 &$\pm$ 1.63e-4 & 5.86 &$\pm$ 0.02 & 2.29 \\
GraphRNN     & 0.315 &$\pm$ 0.006 & 33  &$\pm$ 6.11 & 65.0  &$\pm$ 11.3 & 0.095  &$\pm$ 0.055 & 1.845 &$\pm$ 0.007 & 3.48e-3 &$\pm$ 8.52e-4 & 5.70 &$\pm$ 0.08 & 5.00       \\
VGAE         & 0.509 &$\pm$ 0.002 & 348 &$\pm$ 2.58 & 3731.0 &$\pm$ 22.4 & -0.154  &$\pm$ 0.001 & 2.055 &$\pm$ 0.004 & 1.04e-3 &$\pm$ 2.15e-5 & 5.04 &$\pm$ 0.04 & 6.00 \\
ARVGA        & 0.563 &$\pm$ 0.001 & 239 &$\pm$ 6.12 & 7511.0 &$\pm$ 337.0 & -0.141 &$\pm$ 0.002 & 2.168 &$\pm$ 0.005 & 4.67e-3 &$\pm$ 2.56e-4 & 6.02 &$\pm$ 0.07 & 5.00 \\
\midrule
TVGA     & 0.389 &$\pm$ 0.002 & 152 & $\pm$ 5.84 & 1258.6 &$\pm$ 32.0 & -0.053 &$\pm$ 0.002 &    1.879& $\pm$ 0.002 & 4.26e-3 &$\pm$ 3.29e-4 & 5.42 &$\pm$ 0.03 & 2.00 \\
\bottomrule
\end{tabular}}
\label{tab:coragenerationnostd}
\end{table*}

\begin{table*}[t]
\caption{Statistics for the generated graphs on Citeseer. Results are averaged over 5 runs.}
\centering
\resizebox{0.95\textwidth}{!}{
\begin{tabular}{lrrrrrrrrrrrrrr|c}
\toprule
& \multicolumn{2}{c}{Gini coeff.} & \multicolumn{2}{c}{max degree} &
\multicolumn{2}{c}{triangle count} & \multicolumn{2}{c}{assortativity} &
\multicolumn{2}{c}{power law exp.} & \multicolumn{2}{c}{clustering coeff.} &
\multicolumn{2}{c|}{charac. path len.} & \multicolumn{1}{c}{avg rank} \\
& \multicolumn{1}{c}{avg} & \multicolumn{1}{c}{std} & \multicolumn{1}{c}{avg} &
\multicolumn{1}{c}{std} & \multicolumn{1}{c}{avg} & \multicolumn{1}{c}{std} &
\multicolumn{1}{c}{avg} & \multicolumn{1}{c}{std} & \multicolumn{1}{c}{avg} &
\multicolumn{1}{c}{std} & \multicolumn{1}{c}{avg} & \multicolumn{1}{c}{std} &
\multicolumn{1}{c}{avg} & \multicolumn{1}{c|}{std} & \\
\midrule
original input & 0.428 &       & 99 &      & 1084 &     & 0.008 &       & 2.071 & & 1.30e-2 & & 9.33 & &        \\
\midrule
conf. model & 0.428 &$\pm$0.000 & 99 &$\pm$0.00 & 43.2 & $\pm$8.4 & -0.011 & $\pm$0.010 & 2.071 & $\pm$0.000 & 5.18e-4 & $\pm$1.01e-4 & 5.28 & $\pm$0.04 & 3.43     \\
DC-SBM      & 0.514 & $\pm$0.006 & 90 & $\pm$3.56 & 158.2 & $\pm$11.3 & 0.020 & $\pm$0.006 & 1.957 & $\pm$0.008 & 2.20e-3 & $\pm$1.20e-4 & 5.09 & $\pm$0.03 & 4.29     \\
NetGAN      & 0.365 & $\pm$0.003 & 73.2 & $\pm$4.31 & 592.8 & $\pm$27.5 & -0.043 & $\pm$0.005 & 1.988 & $\pm$0.003 & 1.54e-2 & $\pm$1.60e-3 & 7.55 & $\pm$0.13 & 3.14    \\
GraphRNN    & 0.313 & $\pm$0.005 & 17 & $\pm$1.85 & 89.2 &$\pm$ 7.7    & 0.066  & $\pm$0.015 & 1.964 & $\pm$0.018 & 1.52e-2  & $\pm$1.71e-3 & 7.55 & $\pm$0.25 & 4.43       \\
VGAE        & 0.495 & $\pm$0.003 & 196 & $\pm$4.17 & 4037 & $\pm$114.4 & -0.035 & $\pm$0.006 & 2.221 &$\pm$ 0.006 & 5.73e-3 & $\pm$3.27e-4 & 7.03 & $\pm$0.17 & 5.29     \\
ARVGA       & 0.524 & $\pm$0.004 & 139 & $\pm$8.61 & 6126.8 & $\pm$194.3 & 0.017 & $\pm$0.018 & 2.293 & $\pm$0.012 & 1.99e-2 & $\pm$2.58e-3 & 7.90 & $\pm$0.18 & 4.43     \\

\midrule
TVGA    & 0.428 & $\pm$0.002 & 86 & $\pm$3.07 &  1288.2  & $\pm$40.0 & 0.033  & $\pm$0.007 & 2.068 & $\pm$0.004 & 1.78e-2 & $\pm$1.17e-3 & 6.35 & $\pm$0.06 & 2.71     \\
\bottomrule
\end{tabular}
}
\label{tab:generationciteseer}
\end{table*}

In this experiment,
we train
the model
on an input graph,
and then try to generate similar graphs.
Following \cite{bojchevski2018netgan}, we only experiment
on the Cora and Citeseer data sets. Moreover,
as NetGAN can only generate graphs with the same number of nodes
as the input graph,
we set the numbers of nodes and edges to be generated to be equal to those in the input graph.

The following
baselines
are compared with the proposed model:
(i) configuration model (conf. model) \cite{molloy1995critical}; (ii)
degree-corrected stochastic block model (DC-SBM) \cite{karrer2011stochastic};
(iii) NetGAN\footnote{Here, we use the early stopping scheme
EO-CRITERION. As discussed in
\cite{bojchevski2018netgan},
this gives the user control over graph generation.};
(iv) GraphRNN \cite{you2018graphrnn};
(v) variational graph auto-encoder (VGAE);
and (vi) adversarially regularized variational graph auto-encoder (ARVGA).
The configuration model and DC-SBM are classic
graph generation
algorithms, which
model certain graph statistics directly.
NetGAN generates graphs via accumulating random walks.
GraphRNN generates nodes sequentially.
Besides, note that the configuration model, DC-SBM, NetGAN and GraphRNN do not utilize node features.
Moreover, we do not compare with GraphVAE \cite{simonovsky2018graphvae} and the graph neural network based model in
\cite{li2018learning}, as they can only be used on very small graphs.
\begin{figure}[th!]
\includegraphics[width=.96\columnwidth]{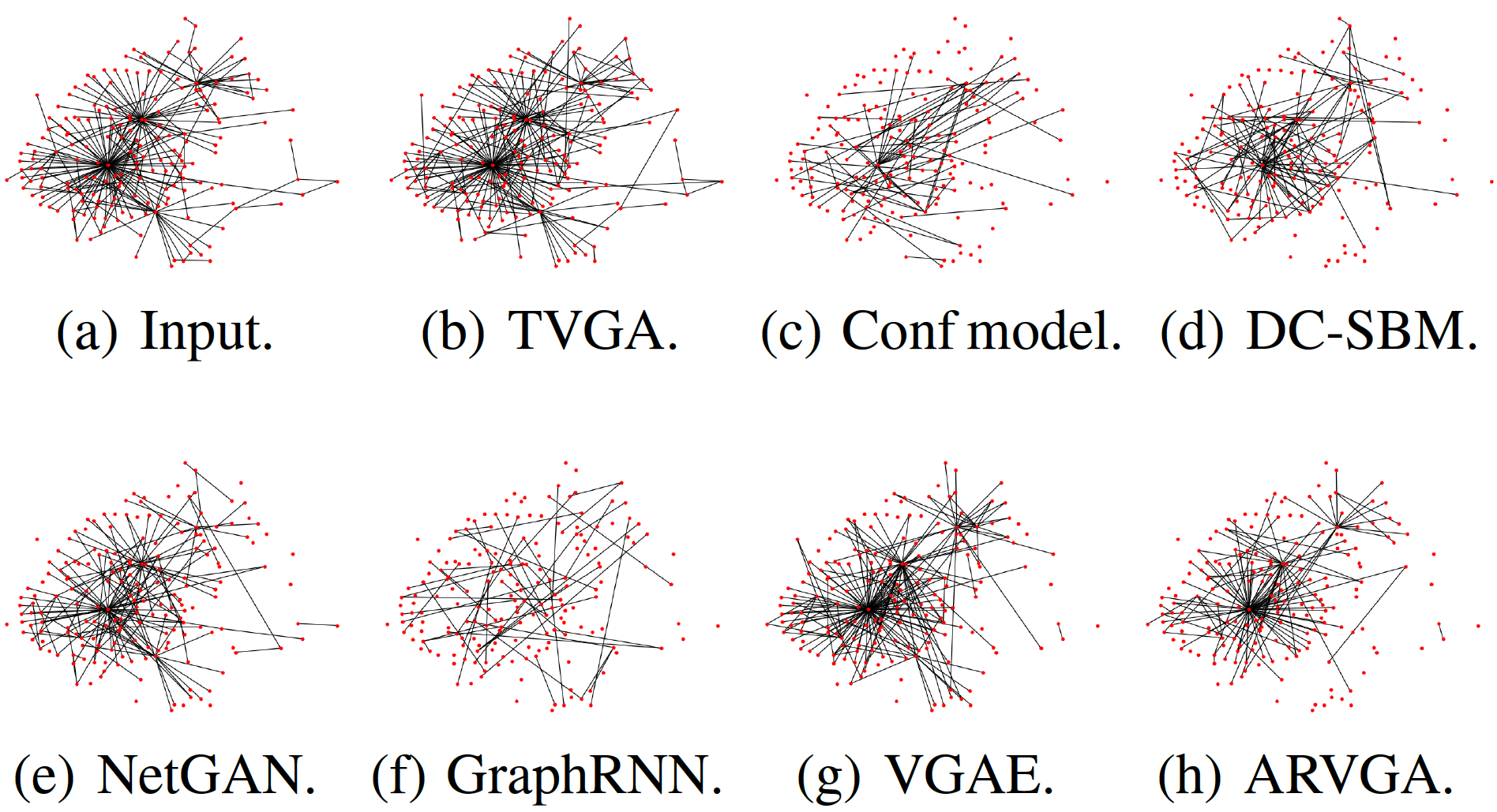}
\vspace{-.1in}
\caption{An original subgraph from Cora, and the corresponding subgraphs generated by various
methods.
}
\label{fig:corageneration}
\end{figure}

As in \cite{bojchevski2018netgan},
performance is evaluated by a number of
statistics
measured on the generated graph. These include
the Gini coefficient,
maximum degree,
number of triangles,
assortativity (i.e., Pearson correlation of degrees of connected nodes),
power law exponent (i.e., exponent of the power law distribution for degrees),
clustering coefficient (as defined in NetGAN),
and
characteristic path length  (i.e., average number of steps along the shortest paths for all node pairs).

Tables~\ref{tab:coragenerationnostd} and \ref{tab:generationciteseer}
show the results
on Cora and Citeseer.
Though the configuration model and
DC-SBM excel at some
metrics
that they directly model (such as ``maximum degree" and ``power law exponent"), they fail
to reproduce others.
GraphRNN does not perform well.
It has to be trained on all breadth-first-search
orderings (which can be in the order of $\mathcal{O}(|V|!)$).
In \cite{you2018graphrnn},
GraphRNN has only been trained on small graphs with a maximum of 500 nodes (while the Cora and
Citeseer data sets here are much larger).
The rightmost columns
in Tables~\ref{tab:coragenerationnostd} and \ref{tab:generationciteseer}
show the average rank of each method over all statistics. TVGA, by utilizing both
pairwise node information and triadic structure, ranks the highest and generates
graphs with similar statistics to the original one
\cite{hamilton2017representation}.

\begin{figure}[t!] \includegraphics[width=.96\columnwidth]{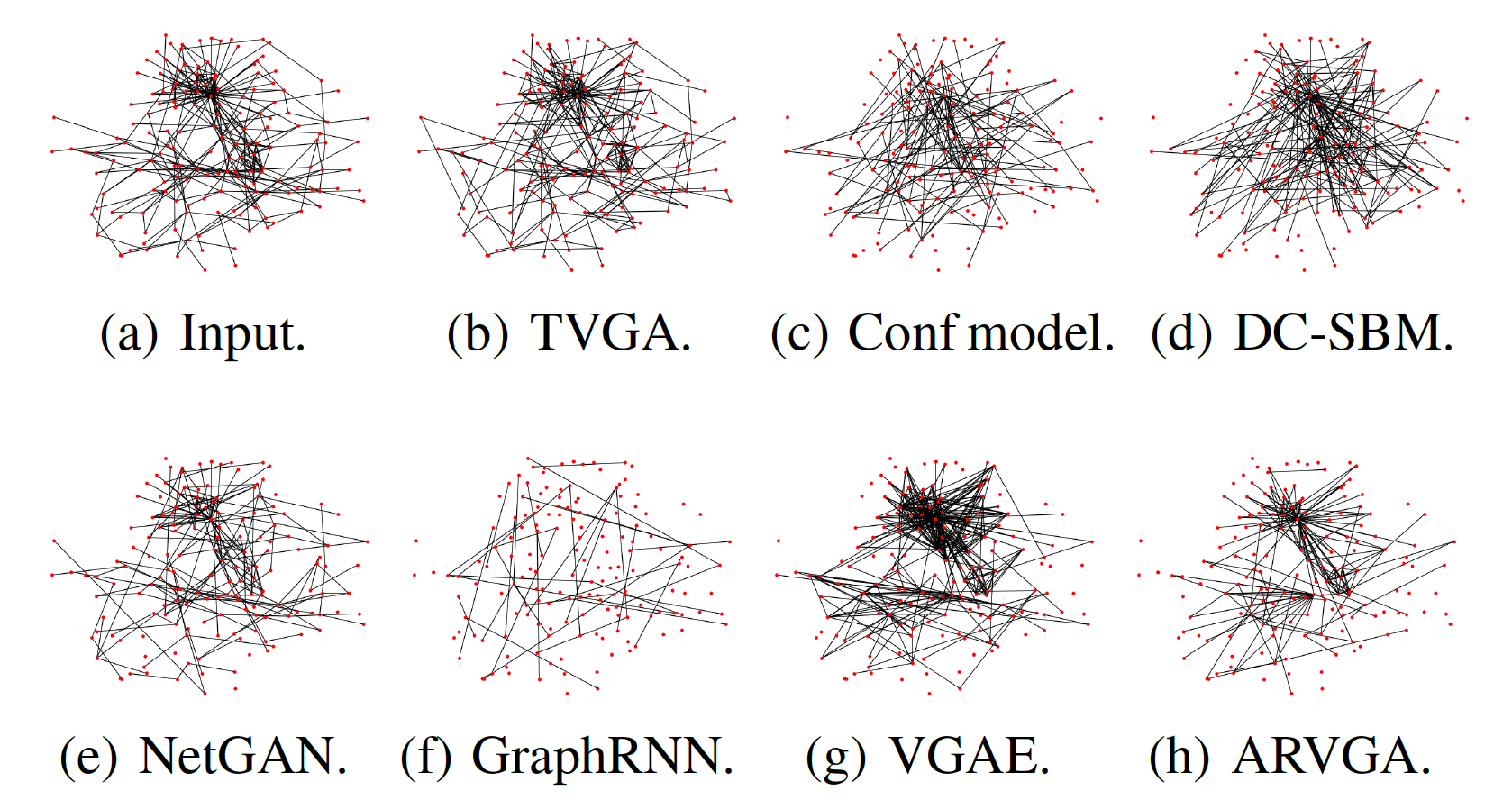}
\vspace{-.1in}
\caption{An original subgraph from Citeseer, and the corresponding subgraphs
generated by various methods.}
\label{fig:citeseergeneration}
\end{figure}

Figures~\ref{fig:corageneration} and \ref{fig:citeseergeneration} show parts of the
generated graphs. As can be seen, the graphs generated by TVGA are more similar to
the original graphs than the others. For example, in Figure~\ref{fig:corageneration}, both the input and TVGA-generated graphs have
three dominant clusters. However, this is not obvious in the other graphs
generated. Similarly, in Figure~\ref{fig:citeseergeneration}, one can
observe three dominant clusters (arranged in a triangle)
in both the input and TVGA-generated graphs, but
not in the others.

\section{Conclusion}

In this paper, we proposed a novel triad decoder that uses the whole local triad information, and is able to model the triadic closure property that is fundamental in real-world networks. It can be readily used in any graph-based auto-encoder. Experimental results show that the proposed decoder, when used with the (variational) graph auto-encoder, outperforms the state-of-the-art on link prediction, node clustering and graph generation tasks.

\section*{Acknowledgement}
This work was supported in part by Huawei PhD Fellowship.

\bibliography{AAAI-ShiH.5197}

\begin{thebibliography}{}

\bibitem[\protect\citeauthoryear{Belanger and
  McCallum}{2016}]{belanger2016structured}
Belanger, D., and McCallum, A.
\newblock 2016.
\newblock Structured prediction energy networks.
\newblock In {\em International Conference on Machine Learning},  983--992.

\bibitem[\protect\citeauthoryear{Bianconi \bgroup et al\mbox.\egroup
  }{2014}]{bianconi2014triadic}
Bianconi, G.; Darst, R.~K.; Iacovacci, J.; and Fortunato, S.
\newblock 2014.
\newblock Triadic closure as a basic generating mechanism of communities in
  complex networks.
\newblock {\em Physical Review E}  042806.

\bibitem[\protect\citeauthoryear{Bojchevski \bgroup et al\mbox.\egroup
  }{2018}]{bojchevski2018netgan}
Bojchevski, A.; Shchur, O.; Z{\"u}gner, D.; and G{\"u}nnemann, S.
\newblock 2018.
\newblock Net{GAN}: Generating graphs via random walks.
\newblock In {\em International Conference on Machine Learning},  609--618.

\bibitem[\protect\citeauthoryear{Bruna \bgroup et al\mbox.\egroup
  }{2014}]{bruna2013spectral}
Bruna, J.; Zaremba, W.; Szlam, A.; and LeCun, Y.
\newblock 2014.
\newblock Spectral networks and locally connected networks on graphs.
\newblock In {\em International Conference on Learning Representations}.

\bibitem[\protect\citeauthoryear{Caplow}{1968}]{caplow1968two}
Caplow, T.
\newblock 1968.
\newblock {\em Two against one: Coalitions in triads}.
\newblock Prentice-Hall.

\bibitem[\protect\citeauthoryear{Globerson \bgroup et al\mbox.\egroup
  }{2015}]{globerson2015hard}
Globerson, A.; Roughgarden, T.; Sontag, D.; and Yildirim, C.
\newblock 2015.
\newblock How hard is inference for structured prediction?
\newblock In {\em International Conference on Machine Learning},  2181--2190.

\bibitem[\protect\citeauthoryear{Goyal and Ferrara}{2018}]{goyal2018graph}
Goyal, P., and Ferrara, E.
\newblock 2018.
\newblock Graph embedding techniques, applications, and performance: A survey.
\newblock {\em Knowledge-Based Systems}  78--94.

\bibitem[\protect\citeauthoryear{Granovetter}{1973}]{granovetter1973strength}
Granovetter, M.
\newblock 1973.
\newblock The strength of weak ties.
\newblock {\em American Journal of Sociology}  1360--1380.

\bibitem[\protect\citeauthoryear{Grover and
  Leskovec}{2016}]{grover2016node2vec}
Grover, A., and Leskovec, J.
\newblock 2016.
\newblock node2vec: Scalable feature learning for networks.
\newblock In {\em International Conference on Knowledge Discovery and Data
  Mining},  855--864.

\bibitem[\protect\citeauthoryear{Hamilton, Ying, and
  Leskovec}{2017}]{hamilton2017representation}
Hamilton, W.; Ying, R.; and Leskovec, J.
\newblock 2017.
\newblock Representation learning on graphs: Methods and applications.
\newblock {\em IEEE Data Engineering Bulletin}  52--74.

\bibitem[\protect\citeauthoryear{He \bgroup et al\mbox.\egroup
  }{2016}]{he2016deep}
He, K.; Zhang, X.; Ren, S.; and Sun, J.
\newblock 2016.
\newblock Deep residual learning for image recognition.
\newblock In {\em International Conference on Computer Vision and Pattern
  Recognition},  770--778.

\bibitem[\protect\citeauthoryear{Karrer and
  Newman}{2011}]{karrer2011stochastic}
Karrer, B., and Newman, M.
\newblock 2011.
\newblock Stochastic blockmodels and community structure in networks.
\newblock {\em Physical Review E} 83(1).

\bibitem[\protect\citeauthoryear{Kingma and Ba}{2014}]{kingma2014adam}
Kingma, D., and Ba, J.
\newblock 2014.
\newblock Adam: A method for stochastic optimization.
\newblock In {\em International Conference on Learning Representations}.

\bibitem[\protect\citeauthoryear{Kingma and Welling}{2014}]{kingma2013auto}
Kingma, D., and Welling, M.
\newblock 2014.
\newblock Auto-encoding variational bayes.
\newblock In {\em International Conference on Learning Representations}.

\bibitem[\protect\citeauthoryear{Kipf and Welling}{2016a}]{kipf2016semi}
Kipf, T., and Welling, M.
\newblock 2016a.
\newblock Semi-supervised classification with graph convolutional networks.
\newblock In {\em International Conference on Learning Representations}.

\bibitem[\protect\citeauthoryear{Kipf and Welling}{2016b}]{kipf2016variational}
Kipf, T., and Welling, M.
\newblock 2016b.
\newblock Variational graph auto-encoders.
\newblock In {\em NIPS Workshop on Bayesian Deep Learning}.

\bibitem[\protect\citeauthoryear{Lawler}{2001}]{lawler2001combinatorial}
Lawler, E.~L.
\newblock 2001.
\newblock {\em Combinatorial optimization: Networks and matroids}.
\newblock Dover.

\bibitem[\protect\citeauthoryear{Li \bgroup et al\mbox.\egroup
  }{2018}]{li2018learning}
Li, Y.; Vinyals, O.; Dyer, C.; Pascanu, R.; and Battaglia, P.
\newblock 2018.
\newblock Learning deep generative models of graphs.
\newblock Preprint arXiv:1803.03324.

\bibitem[\protect\citeauthoryear{Molloy and Reed}{1995}]{molloy1995critical}
Molloy, M., and Reed, B.
\newblock 1995.
\newblock A critical point for random graphs with a given degree sequence.
\newblock {\em Random Structures \& Algorithms} 6(2-3):161--180.

\bibitem[\protect\citeauthoryear{Munkres}{1957}]{munkres1957algorithms}
Munkres, J.
\newblock 1957.
\newblock Algorithms for the assignment and transportation problems.
\newblock {\em Journal of the Society for Industrial and Applied Mathematics}
  32--38.

\bibitem[\protect\citeauthoryear{Pan \bgroup et al\mbox.\egroup
  }{2018}]{pan2018adversarially}
Pan, S.; Hu, R.; Long, G.; Jiang, J.; Yao, L.; and Zhang, C.
\newblock 2018.
\newblock Adversarially regularized graph autoencoder for graph embedding.
\newblock In {\em International Joint Conferences on Artificial Intelligence}.

\bibitem[\protect\citeauthoryear{Sen \bgroup et al\mbox.\egroup
  }{2008}]{sen2008collective}
Sen, P.; Namata, G.; Bilgic, M.; Getoor, L.; Galligher, B.; and Eliassi-Rad, T.
\newblock 2008.
\newblock Collective classification in network data.
\newblock {\em AI Magazine} ~93.

\bibitem[\protect\citeauthoryear{Simmel}{1908}]{simmel1908sociology}
Simmel, G.
\newblock 1908.
\newblock {\em Sociology: Investigations on the forms of sociation}.

\bibitem[\protect\citeauthoryear{Simonovsky and
  Komodakis}{2018}]{simonovsky2018graphvae}
Simonovsky, M., and Komodakis, N.
\newblock 2018.
\newblock Graph{VAE}: Towards generation of small graphs using variational
  autoencoders.
\newblock In {\em International Conference on Artificial Neural Networks},
  412--422.

\bibitem[\protect\citeauthoryear{Wang \bgroup et al\mbox.\egroup
  }{2017}]{wang2017mgae}
Wang, C.; Pan, S.; Long, G.; Zhu, X.; and Jiang, J.
\newblock 2017.
\newblock {MGAE}: Marginalized graph autoencoder for graph clustering.
\newblock In {\em International Conference on Information and Knowledge
  Management},  889--898.

\bibitem[\protect\citeauthoryear{Wang, Chen, and Li}{2017}]{wang2017predictive}
Wang, Z.; Chen, C.; and Li, W.
\newblock 2017.
\newblock Predictive network representation learning for link prediction.
\newblock In {\em International ACM SIGIR Conference on Research and
  Development in Information Retrieval},  969--972.

\bibitem[\protect\citeauthoryear{Wasserman and
  Faust}{1994}]{wasserman1994social}
Wasserman, S., and Faust, K.
\newblock 1994.
\newblock {\em Social network analysis: Methods and applications}.
\newblock Cambridge University Press.

\bibitem[\protect\citeauthoryear{Xia \bgroup et al\mbox.\egroup
  }{2014}]{xia2014robust}
Xia, R.; Pan, Y.; Du, L.; and Yin, J.
\newblock 2014.
\newblock Robust multi-view spectral clustering via low-rank and sparse
  decomposition.
\newblock In {\em AAAI Conference on Artificial Intelligence},  2149--2155.

\bibitem[\protect\citeauthoryear{Yang \bgroup et al\mbox.\egroup
  }{2015}]{yang2015network}
Yang, C.; Liu, Z.; Zhao, D.; Sun, M.; and Chang, E.
\newblock 2015.
\newblock Network representation learning with rich text information.
\newblock In {\em International Joint Conferences on Artificial Intelligence},
  2111--2117.

\bibitem[\protect\citeauthoryear{You \bgroup et al\mbox.\egroup
  }{2018}]{you2018graphrnn}
You, J.; Ying, R.; Ren, X.; Hamilton, W.; and Leskovec, J.
\newblock 2018.
\newblock Graph{RNN}: Generating realistic graphs with deep auto-regressive
  models.
\newblock In {\em International Conference on Machine Learning},  5694--5703.

\bibitem[\protect\citeauthoryear{Zhang and Chen}{2018}]{zhang2018link}
Zhang, M., and Chen, Y.
\newblock 2018.
\newblock Link prediction based on graph neural networks.
\newblock In {\em Advances in Neural Information Processing Systems},
  5165--5175.

\bibitem[\protect\citeauthoryear{Zhang \bgroup et al\mbox.\egroup
  }{2018}]{zhang2018network}
Zhang, D.; Yin, J.; Zhu, X.; and Zhang, C.
\newblock 2018.
\newblock Network representation learning: A survey.
\newblock {\em IEEE Transactions on Big Data}.

\bibitem[\protect\citeauthoryear{Zhou \bgroup et al\mbox.\egroup
  }{2018}]{zhou2018dynamic}
Zhou, L.; Yang, Y.; Ren, X.; Wu, F.; and Zhuang, Y.
\newblock 2018.
\newblock Dynamic network embedding by modeling triadic closure process.
\newblock In {\em AAAI Conference on Artificial Intelligence}.

\end{thebibliography}
\bibliographystyle{aaai}

\end{document}